\documentclass[letterpaper,10pt,conference]{ieeeconf}
\IEEEoverridecommandlockouts       
\usepackage{amssymb}
\usepackage{amsmath}
\usepackage[english]{babel}
\usepackage{float}
\usepackage{graphicx}
\usepackage{hyperref}
\usepackage{mathtools}

\usepackage{filecontents}
\usepackage[noadjust]{cite}
\usepackage[font=footnotesize,labelfont=footnotesize]{subcaption}
\usepackage[font=footnotesize,labelfont=footnotesize]{caption}
\usepackage{comment}
\usepackage{authblk}
\usepackage{multirow}
\usepackage{xcolor}
\usepackage{algorithm}
\usepackage{algorithmic}
\usepackage{xspace}

\DeclareMathOperator*{\argmax}{arg\,max}

\definecolor{darkblue}{rgb}{0.15,0.15,0.55}
\definecolor{lightgrey}{rgb}{0.75,0.75,0.75}

\providecommand{\codecomment}[1]{\textcolor{lightgrey}{\dotfill}\textcolor{darkblue}{//\,\textrm{#1}}}

\newcommand{\nphard}{$\mathcal{NP}$-hard\xspace}

\newcommand{\thm}{\noindent \textbf{Theorem}\xspace}
\newcommand{\lem}{\noindent \textbf{Lemma}\xspace}

\newcommand{\pf}{\noindent \textbf{Proof}\xspace}

\newcommand{\qed}{\hfill $\square$}

\title{\LARGE \bf Fast and resilient manipulation planning for target retrieval in clutter \vspace{-3pt}}
\author{Changjoo Nam, Jinhwi Lee, Sang Hun Cheong, Brian Y. Cho, and ChangHwan Kim$^*$
\thanks{This work was supported by the Technology Innovation Program and Industrial Strategic Technology Development Program (10077538, Development of manipulation technologies in social contexts for human-care service robots). The authors are with Korea Institute of Science and Technology.
$^*$Corresponding author: {\tt\small ckim@kist.re.kr}. }}

\begin{document}
\maketitle

\begin{abstract}
This paper presents a task and motion planning (TAMP) framework for a robotic manipulator in order to retrieve a target object from clutter. We consider a configuration of objects in a confined space with a high density so no collision-free path to the target exists. The robot must relocate some objects to retrieve the target without collisions.  For fast completion of object rearrangement, the robot aims to optimize the number of pick-and-place actions which often determines the efficiency of a TAMP framework.

We propose a task planner incorporating motion planning to generate executable plans which aims to minimize the number of pick-and-place actions. In addition to fully known and static environments, our method can deal with uncertain and dynamic situations incurred by occluded views. Our method is shown to reduce the number of pick-and-place actions compared to baseline methods (e.g., at least 28.0\% of reduction in a known static environment with 20 objects). 
\end{abstract}

\section{Introduction}
\vspace{-2pt}

\begin{figure}[t!]
    \captionsetup{skip=0pt}
    \centering
	\includegraphics[width=0.42\textwidth]{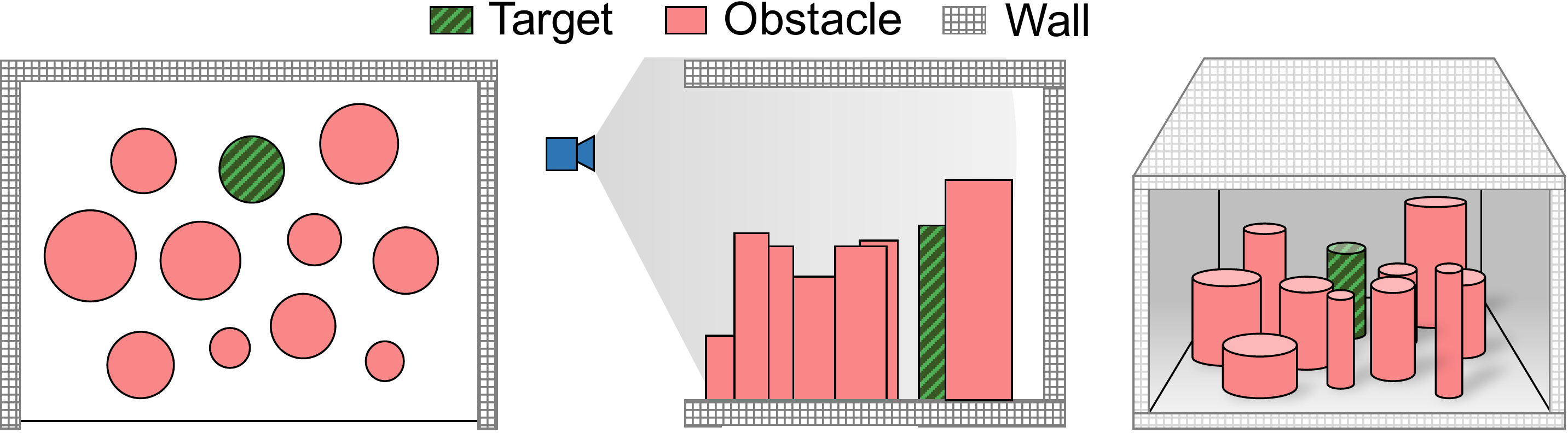}
    \caption{Objects in dense clutter. Objects could be occluded and inaccessible.}
    \label{fig:example}
    \vspace{-20pt}
\end{figure}

Retrieving a target object from clutter using a robotic manipulator has long been considered as an important and practical task. Robots will perform such tasks in cluttered and confined spaces frequently in our home or workplace (e.g., shelves in fridges or cabinets) as illustrated in Fig.~\ref{fig:example}. If objects are populated densely and overhand grasps are not allowed, a subset of the objects should be relocated to secure a collision-free path for the manipulator to retrieve the target. 

In manipulation planning, task planning focuses on generating high-level discrete actions while motion planning finds a sequence of robot configurations which result in continuous motions. Recently, tight coupling of task and motion planning (TAMP) has shown successful achievements~\cite{dantam2018task,garrett2018ffrob,srivastava2014combined,toussaint2015logic} by generating symbolic task plans that are executable. However, TAMP frameworks could be inefficient for manipulation tasks in dense clutter. Planning robot motions in dense clutter could fail frequently but geometric motion planners cannot provide the cause of failure in the form of logical constraints. Thus, the task planners should iterate over symbolic plans if motion planning fails~\cite{lagriffoul2016combining}.

In this work, we propose a planning method taking the TAMP approach where the geometric task planner is specified for object rearrangement in dense clutter. Our task planner aims to generate a sequence of feasible pick-and-place actions to relocate objects until a target object becomes reachable. In order to establish a task plan that has valid continuous motions, the actions should be examined by a motion planner. However, determining the sequence is difficult as manipulation planning among movable obstacles (MAMO) has shown to be \nphard even in fully known and static environments~\cite{stilman2007manipulation}. Cluttered and confined environments make the problem harder.

We develop an efficient task and motion planning algorithm which is resilient to motion planning failures. The algorithm aims to minimize the number of pick-and-place actions which often determines the efficiency of MAMO solvers~\cite{han2018complexity,han2018efficient}.
We begin from considering a fully known environment. Then, we take into account uncertain and dynamic situations arising from occlusions. Since the focus of this work is on generating valid task plans in conjunction with motion planning,  other issues like grasp planning and precision control are not considered. Non-prehensile actions for rearranging objects~\cite{dogar2012planning,yuan2018rearrangement} are not used in this work but could be a part of our future work.

The following are contribution of this work:\vspace{-3pt}
\begin{itemize}
    \item We propose a polynomial-time algorithm that constructs a concise data structure that describes feasible discrete manipulation actions by processing geometric information of objects and the robot end-effector (Sec.~\ref{sec:tgraph}). 
    \item Using the data structure, we propose a task planning algorithm that determines a sequence of pick-and-place actions which have continuous motions satisfying kinematic constraints of the manipulator. The proposed method can deal with uncertainties arising from occlusions (Sec.~\ref{sec:reloc}).
    \item We provide proofs for the time complexity and completeness of some proposed algorithms (Sec.~\ref{sec:analysis}).
    \item We show results from extensive simulations in various scenarios (Sec.~\ref{sec:exp}).
\end{itemize}

\section{Related Work}
\label{sec:related}
\vspace{-2pt}
The work presented in~\cite{dogar2012planning} proposes a planning framework to grasp a target in cluttered and known environments. It removes obstacles that are in the shortest path of the end-effector to the target. Although this method finds the distance-optimal path, some obstacles could be removed unnecessarily since it does not optimize the number of objects to be removed. Other works, such as \cite{srivastava2014combined,haustein2015kinodynamic,moll2018randomized}, also do not minimize the number of manipulation actions but mainly concern about validity of the plan.

Some recent work considers partially known environments. The algorithm proposed in~\cite{dogar2014object} computes a sequence of objects to be removed while minimizing the expected time to find a hidden target. The strength of this work is the mathematical formalization of the search and grasp planning problem. However, the algorithm shows exponential running time so may not be practically useful in environments with densely packed objects (planning takes longer than 25\,sec for five objects). 
Another work~\cite{lin2015planning} finds a sequence of actions of a mobile manipulator that minimizes the expected time to reveal all possible hidden target poses. This work defines admissible costs for its A$^*$ search, but planning takes long time owing to the high branching factor of the search (e.g., 40\,sec with five objects).
 
In \cite{krontiris2015dealing}, the authors present methods dealing with object rearrangement. Although the suggested methods show high success rates in a confined space where overhand grasps are not allowed, they do not scale to the number of objects. \cite{vega2016asymptotically} propose an asymptotically optimal algorithm for rearrangement and manipulation planning. 
However, the search in a graph structure takes exponential time (e.g., few seconds for an instance with only two objects). 
\cite{han2018complexity} and \cite{huang2019large} consider object rearrangement on a tabletop. In both, scalable algorithms are proposed which can handle hundreds objects. However, their problems allow overhand grasps so motion planning is relatively easy.

Our own work~\cite{lee2019efficient} proposes a fast algorithm for object relocation in clutter where overhead grasps are not allowed. 
However, it is a local planner so does not set out to achieve the global optimum. Also, it does not consider motion planning of the whole manipulator and partially known environments. In this work, we aim to develop a fast algorithm that optimizes the number of manipulation actions while considering robot kinematic constraints and partially known environments caused by occlusions.

\section{Problem Formulation}
\label{sec:def}
\vspace{-2pt}
Target retrieval from clutter requires several different processes such as perception, task planning, motion planning, and grasping. We focus on the relocation task and motion planning.
The problem of finding a path in a configuration of movable objects has an exponentially large search space in the number of objects. A simplified version of the problem with only one object is shown to be \nphard~\cite{wilfong1991motion,stilman2008planning} even in a perfectly known environment. We consider known environments as well as uncertainties incurred by occlusions. 

Major assumptions: (i) No collision-free path exists for the end-effector without relocating some objects. The spaces to place relocated objects predetermined to be outside the workspace.\footnote{Relocating objects inside the confined space is done in our sibling paper~\cite{cheong2020where}.} (ii) Overhand grasps are not allowed (e.g., the top is blocked by shelves). (iii) Objects are modeled by 3D cylinders (which could have different radii) so the objects can be grasped from any direction. 

\subsection{Problem definition}
\begin{figure}[h!]
\vspace{-8pt}
  \begin{center}
    \begin{minipage}{0.4\linewidth}
      \includegraphics[scale=0.38]{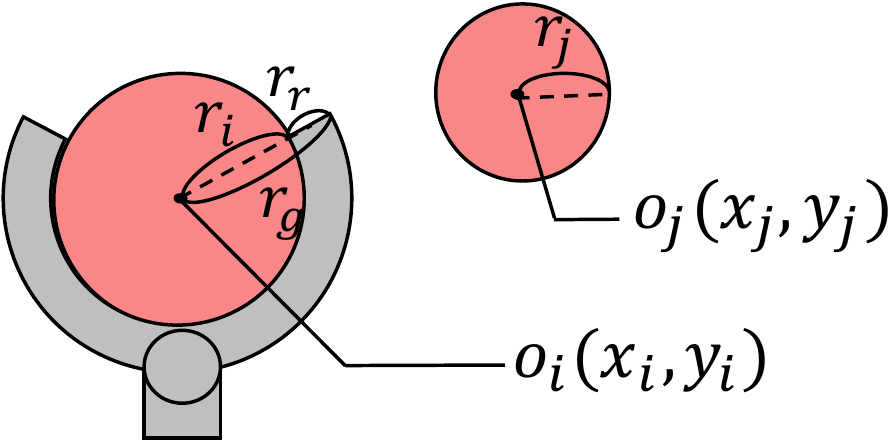}
    \end{minipage}\quad
    \begin{minipage}{0.4\linewidth}
	\caption{Object $i$ at $(x_i, y_i)$ with a radius $r_i$. If $o_i$ grasped, the size of the end-effector $r_r$ is added to $r_i$.}\label{fig:geo}
    \end{minipage}
    \vspace{-10pt}
  \end{center}
  \vspace{-12pt}
\end{figure}

\vspace{-2pt}
Our goal is to complete the target retrieval task quickly so our objective value is the number of pick-and-place actions (the number of relocated objects equivalently). Suppose that an environment is with obstacles $o_i$ for $i = 1, \cdots, N-1 \in \mathbb{Z}^{+}$ and a target $o_t$ (so total $N$ objects). The centroid, radius, and height of object $i$ is described by $(x_i, y_i)$, $r_i$, and $h_i$, respectively. The set $\mathcal{O}$ includes all objects so $\mathcal{O} = \{o_1, \cdots, o_{N-1}, o_t\}$. Let $\mathcal{O}_R \subset \mathcal{O}$ be the sequence of objects to be relocated (including the target) where $|\mathcal{O}_R| = k \le N$. 
The home pose of the robot end-effector is described by $p_R$. The thickness of the end-effector is $r_r$. If it grasps an object whose radius is $r_i$, the radius of the end-effector grasping the object is $r_g = r_i + r_r$. Fig.~\ref{fig:geo} shows the geometry of objects and the end-effector. The camera is fixed at $(x_c, y_c, h_c)$.

A mathematical definition of the problem is to find $\mathcal{O}_R$ that minimizes $k$. The solution sequence $\mathcal{O}_R$ lists objects in the order in which they should be removed.


\subsection{Dynamic and uncertain situations owing to occlusion}
\label{sec:cases}
\begin{figure}[h!]
\vspace{-7pt}
    \captionsetup{skip=0pt}
    \centering
    \begin{subfigure}{0.1335\textwidth}
    \captionsetup{skip=0pt}
	    \includegraphics[width=\textwidth]{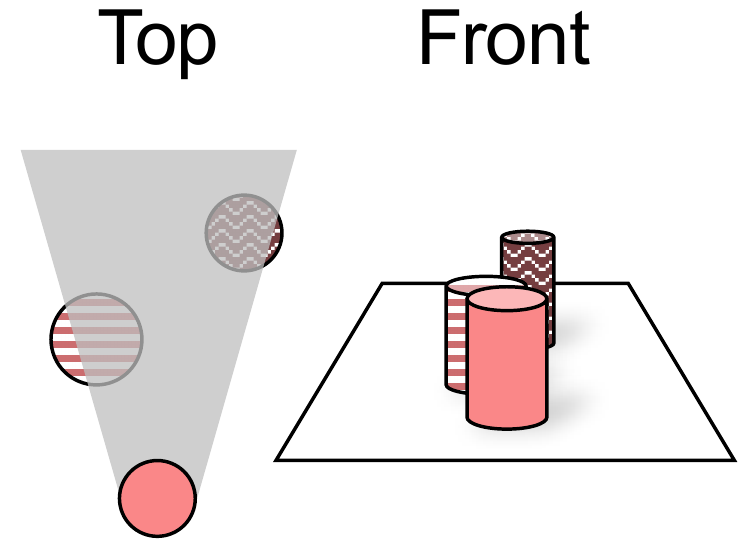}
        \caption{Occluded objects}
        \label{fig:o_obj}
    \end{subfigure}%
    \begin{subfigure}{0.215\textwidth}
    \captionsetup{skip=0pt}
	    \includegraphics[width=\textwidth]{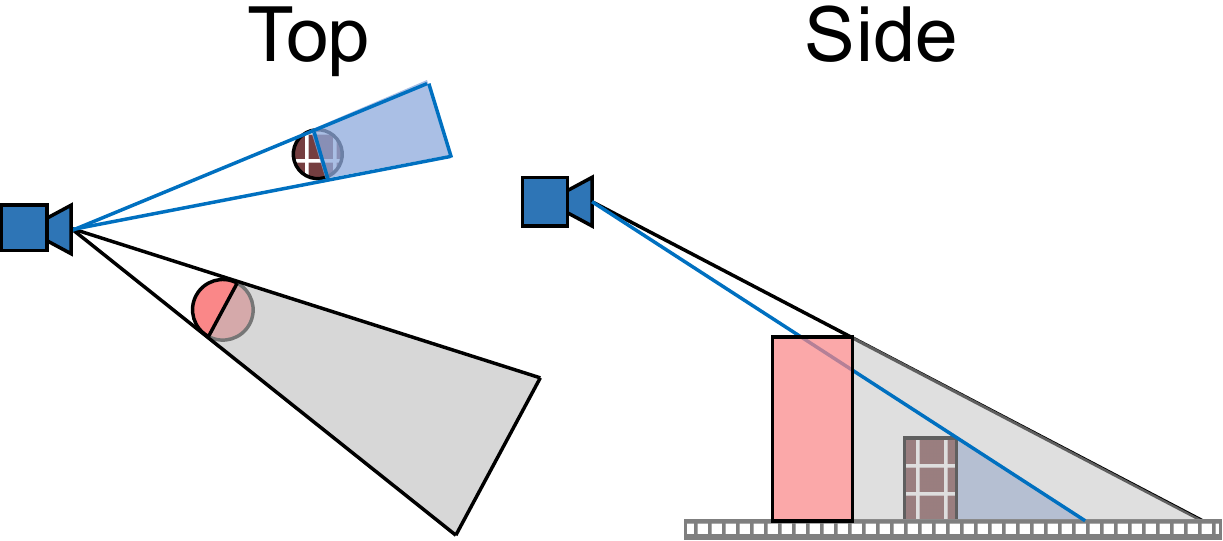}
        \caption{Occluded volume (shades)}
        \label{fig:o_vol}
    \end{subfigure}
    \begin{subfigure}{0.13\textwidth}
        \captionsetup{skip=0pt}
	    \includegraphics[width=\textwidth]{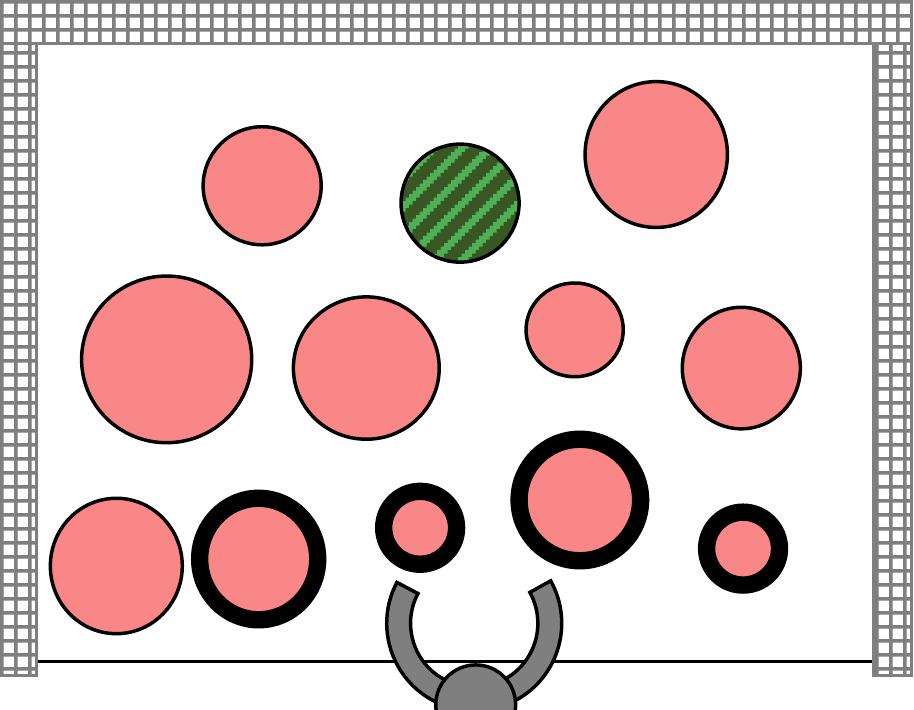}
        \caption{Accessible objects (bold outlines)}
        \label{fig:acc}
  \end{subfigure}
    \caption{Concepts related to occlusion in clutter.}
  \label{fig:defs}
\vspace{-8pt}
\end{figure}

\vspace{-2pt}
Objects could occlude each other in dense clutter. We need to consider different situations occurring from occlusions so define relevant concepts. An object is \textit{occluded} if it is partially visible to the robot. \textit{Occluded volume} quantifies the space occluded by objects (Fig.~\ref{fig:o_vol}). An object is \textit{accessible} if it can be grasped by the end-effector without relocating any objects. The set $\mathcal{O}_A \subset O$ includes all accessible objects. 
Fig.~\ref{fig:acc} shows four accessible objects (bold outlines). The leftmost object at the bottom is not accessible since there is no space for the end-effector to wedge its fingers into.

\begin{figure}[h!]
\vspace{-7pt}
    \captionsetup{skip=0pt}
    \centering
	\includegraphics[width=0.38\textwidth]{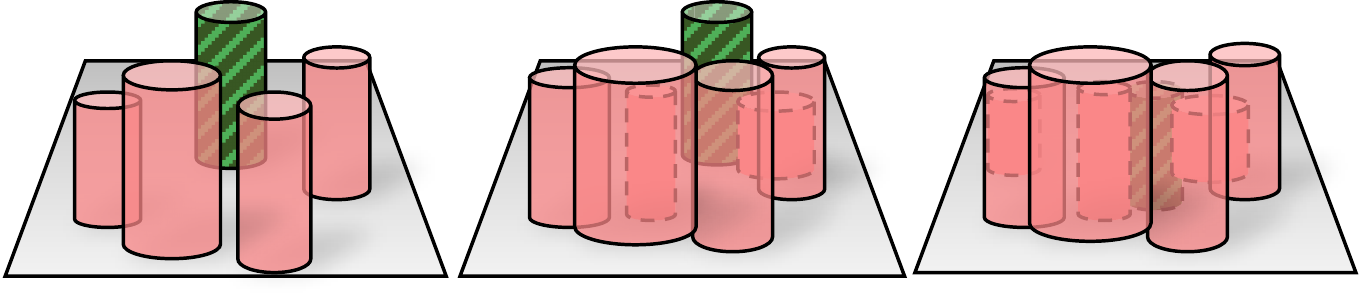}
    \caption{(L) Case I: All objects including the target (green stripes) are known. (C) Case II: Some objects excluding the target are unknown (dotted outlines). (R) Case III: Some objects including the target are unknown.}
  \label{fig:uncertain}
\vspace{-13pt}
\end{figure}

We consider three cases in Fig.~\ref{fig:uncertain}. \underline{Case I:} Known geometry of $\mathcal{O}$ and detected target.
\underline{Case II:} Partially known geometry of $\mathcal{O}$ and detected target. Some hidden obstacles appear dynamically during execution. 
\underline{Case III:} Partially known geometry of $\mathcal{O}$ and undetected target. Some hidden objects  appear dynamically.

\section{Task and motion planning for object retrieval from clutter}
\label{sec:alg}
\vspace{-2pt}
Our task and motion planning framework consists of graph construction followed by task planning combined with motion planning. The graph represents traversability of objects considering the size of the robot hand. The task and motion planner finds a sequence of objects to be relocated using the graph while considering the robot kinematic constraints. 

\subsection{Traversability graph}
\label{sec:tgraph}
\begin{figure}[h!]
\vspace{-5pt}
    \captionsetup{skip=0pt}
    \centering
   \begin{subfigure}{0.095\textwidth}
   \captionsetup{skip=0pt}
	\includegraphics[width=\textwidth]{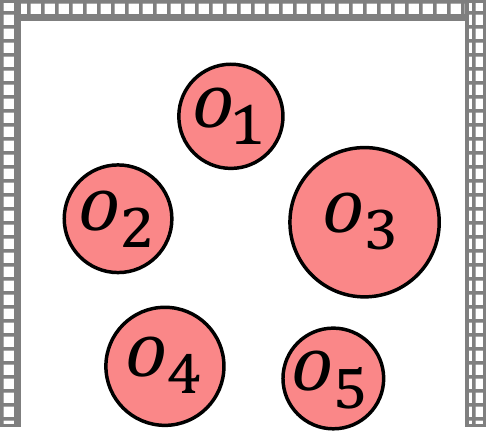}
	\caption{}
    \label{fig:idea1}
  \end{subfigure}%
  \begin{subfigure}{0.095\textwidth}
  \captionsetup{skip=0pt}
	\includegraphics[width=\textwidth]{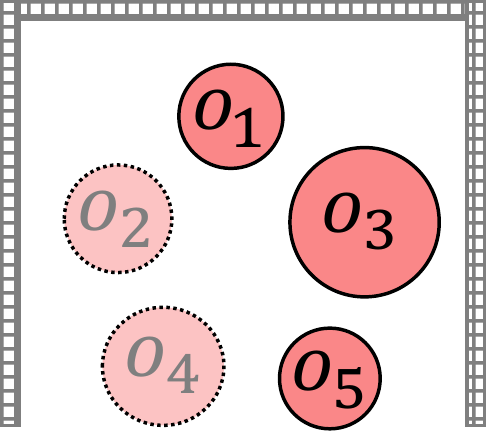}
	\caption{}
    \label{fig:idea2}
  \end{subfigure}%
  \begin{subfigure}{0.094\textwidth}
  \captionsetup{skip=0pt}
	\includegraphics[width=\textwidth]{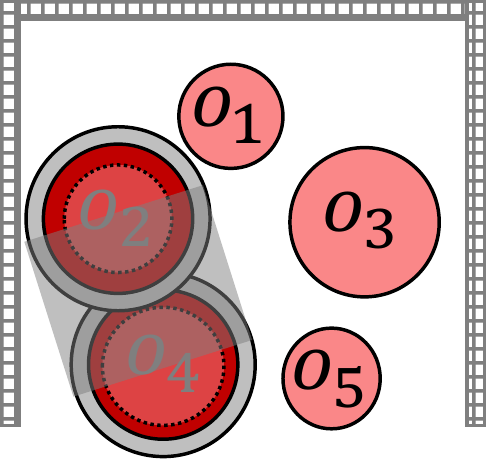}
	\caption{}
    \label{fig:idea3}
  \end{subfigure}%
 \begin{subfigure}{0.095\textwidth}
   \captionsetup{skip=0pt}
	\includegraphics[width=\textwidth]{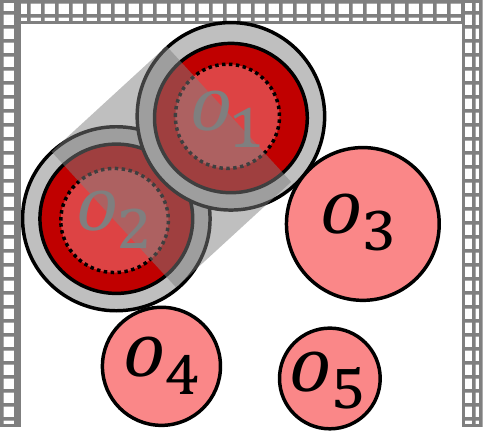}
	\caption{}
    \label{fig:idea4}
  \end{subfigure}%
 \begin{subfigure}{0.094\textwidth}
   \captionsetup{skip=0pt}
	\includegraphics[width=\textwidth]{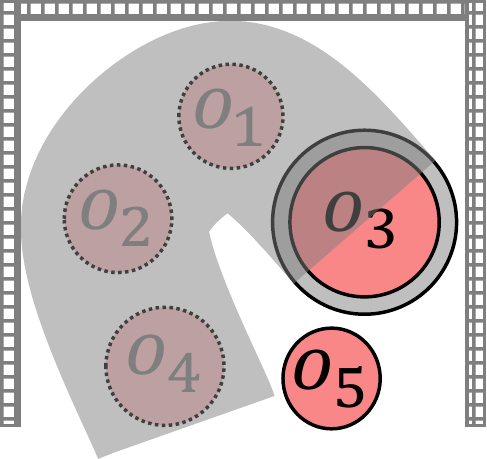}
	\caption{}
    \label{fig:idea5}
  \end{subfigure}
  \caption{A path for target retrieval. (a) An initial configuration. (b) Suppose that $o_2$ and $o_4$ do not exist. (c) If the end-effector grasping the largest object $o_3$ (the gray ring adds the end-effector size) can move between the poses of $o_2$ and $o_4$ without collision, a path exists between the two poses. (d) The same applies to the path between $o_1$ and $o_2$. (e) An example trajectory that $o_3$ can be retrieved from the clutter if the objects on the path are removed.}
  \label{fig:idea}
\vspace{-12pt}
\end{figure}

\begin{figure}[h!]
\vspace{-5pt}
  \begin{center}
    \begin{minipage}{0.4\linewidth}
      \includegraphics[scale=0.33]{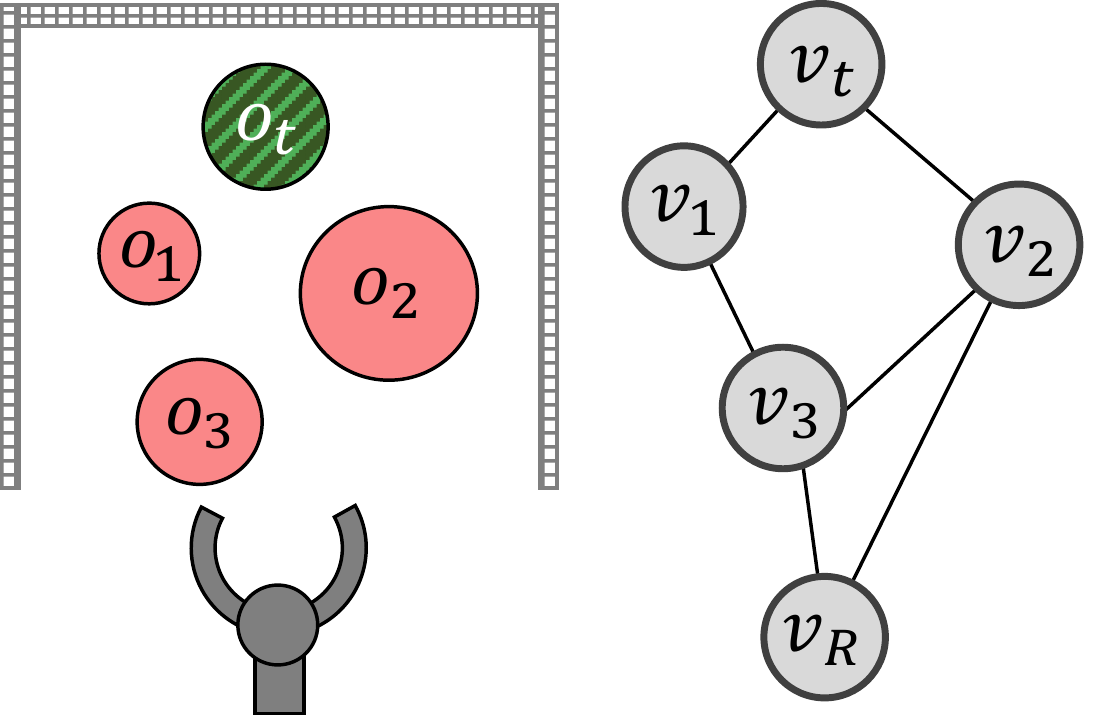}
    \end{minipage}\quad
    \begin{minipage}{0.5\linewidth}
	\caption{An example of T-graph construction. The largest object $o_2$ cannot move between $(x_1, y_1)$ and $(x_2, y_2)$, so nodes $v_1$ and $v_2$ are not connected. Similarly, $v_t$ and $v_3$ are not connected. The end-effector can access $o_2$ and $o_3$ so $v_R$ has edges incident to $v_2$ and $v_3$.}
  \label{fig:graph_construction}
    \end{minipage}
    \vspace{-10pt}
  \end{center}
  \vspace{-7pt}
\end{figure}

We construct a \textit{traversability graph} (T-graph) representing movable paths of objects in clutter. An edge between a pair of nodes means a collision-free path of the end-effector to move any object between the two poses represented by the nodes. The basic idea is illustrated in Figs.~\ref{fig:idea} and \ref{fig:graph_construction}. Any object in $\mathcal{O}$ can move between two poses if a path exists between the two corresponding nodes in the graph and the objects in the path are cleared. For example, $\mathcal{V}_R = \{v_R, v_2, v_t \}$ in Fig.~\ref{fig:graph_construction} is the shortest path from $v_R$ to $v_t$ so the end-effector can retrieve $o_t$ if $o_2$ is removed.

The T-graph construction is described in Alg.~\ref{alg:graph}, which constructs an unweighted and undirected graph $G(\mathcal{V}, \mathcal{E})$ from $\mathcal{O}$ where $\mathcal{V}$ and $\mathcal{E}$ are the sets of nodes and edges, respectively. Nodes represent the objects in $\mathcal{O}$ so $\mathcal{V} = \{v_1, \cdots, v_t, v_R\}$ where $v_t$ and $v_R$ are the target and robot node, respectively (line 2). For every pair of nodes $v_i$ and $v_j$ where $i \neq j$, an edge $(i, j) \in \mathcal{E}$ is connected if any object grasped by the end-effector can move between the poses of $o_i$ and $o_j$ without a collision (lines 5--11). We use $r_g = r_{\mbox{\scriptsize max}} + r_r + r_s$ for collision checking where $r_{\mbox{\scriptsize max}}$ is the radius of the largest object, $r_r$ is the end-effector size $r_r$, and $r_s$ is the safety margin (line 4). Immovable objects (e.g., walls) are considered during collision checking. For collision checking, we employ the modified VFH+~\cite{lee2019efficient}, but any available one can replace. 

The T-graph can be computed efficiently by considering the size of objects and the end-effector only. It screens out infeasible pick-and-place actions before motion planning considering the kinematic constraints of the robot arm. Thus, unnecessary computation for motion planning can be reduced in the following relocation planning. 

\vspace{-4pt}
\begin{algorithm}
\caption{\textsc{GenGraph}}
\label{alg:graph}
{\footnotesize
\vspace{.01in}
\textbf{Input:}
object geometry $\mathcal{O}$, workspace $\mathcal{W}$, robot position $p_R$, robot size $r_r$, safety margin $r_s$\\
\textbf{Output:}
an unweighted undirected graph $G = (\mathcal{V}, \mathcal{E})$\\
 
\vspace{-.1in}
1 \ \ $N = |O| - 1$ \codecomment{$N$ counts obstcales only}\\
2 \ \ $\mathcal{V} = \{v_1, \cdots, v_t, v_R\}$ \codecomment{nodes of all objects and the end-effector}\\
3 \ \ $\mathcal{E} = \emptyset$ \\
4 \ \ $r_g = r_{\mbox{\scriptsize max}} + r_r + r_s$\codecomment{$r_{\mbox{\scriptsize max}}$ is the radius of the largest object}\\
5 \ \ \textbf{for each} $v_i \in \mathcal{V}$ \\
6 \ \ \quad $\mathcal{V}^\prime = \mathcal{V} \setminus v_i$ \codecomment{no self-loop considered so $i \neq j$} \\ 
7 \ \ \quad \textbf{for each} $v_j \in \mathcal{V}^\prime$ \\
8 \ \ \quad \quad \textbf{if} $\sim$\textsc{isCollision}$(r_g, o_i, o_j, \mathcal{W})$ \codecomment{true if there is a collision for}\\ \codecomment{moving the largest object between $o_i(x_i, y_i)$ and $o_j(x_j, y_j)$}\\ \codecomment{in the presence of immovable objects in $\mathcal{W}$} \\
9 \ \ \quad \quad \quad $\mathcal{E} \leftarrow \mathcal{E} \cup (i, j)$ \codecomment{add an edge to $\mathcal{E}$} \\
10 \ \quad \textbf{end for}\\
11 \ \textbf{end for}\\
12 \ \textbf{return} $G(\mathcal{V}, \mathcal{E})$
}\end{algorithm}

\subsection{Relocation planning}
\label{sec:reloc}
\vspace{-2pt}
Alg.~\ref{alg:path} finds the minimum-hop path on a T-graph $G$, which is $\mathcal{V}_R$ (line~1) from the robot node $v_R$ to the target $v_t$. The path represents a sequence of objects to be relocated $\mathcal{O}_R$. Path finding on $G$ is implemented using Breadth first search (BFS)~\cite{cormen2009introduction}. If there are multiple paths that have the same number of hops, we compare their Euclidean distances to break the ties (lines~2--4). This algorithm does not consider the kinematic constraints of the manipulator. Thus, Alg.~\ref{alg:base} combines motion planning with task planning where motion planning failures are handled in an online manner to find an alternative.

Initially, a task plan is found offline by Algs.~\ref{alg:graph}~and~\ref{alg:path} (lines~1--2). For each pick-and-place action for object $o \in \mathcal{O}_R$, motion planning is performed (lines~4--5) for both pick and place actions. If feasible motions are found, $o$ is removed from the scene (line~7). After $o$ is removed, the robot senses the scene to reflect the changes in traversability accordingly (lines~8--9). If no motion is found, the edge $(v_R, v)$ (representing the path from the robot to $o$) is removed from $G$ (line~11). If one or more new objects are found, they are included in the graph to update the task plan (lines~13--15). The whole procedure iterates until $\mathcal{O}_R$ becomes empty. If the robot does not have feasible motions for picking and placing any of the objects, the algorithm fails (lines~17--19).

\begin{algorithm}
\caption{\textsc{RelocPath}}
\label{alg:path}
{\footnotesize
\vspace{.01in}
\textbf{Input:}
graph $G(\mathcal{V}, \mathcal{E})$, target and robot poses $o_t, p_R$\\
\textbf{Output:}
a sequence of objects to be relocated $\mathcal{O}_R$\\
 
\vspace{-.1in}
1 \ $\mathcal{V}_R = \mbox{\textsc{MinHopPath}}(G, v_R, v_t)$ \codecomment{find a min-hop path $\mathcal{V}_R$ with}\\ \codecomment{$k$ nodes from the robot $v_R$ to the target $v_t$}\\ \codecomment{nodes $v_t$ and $v_R$ represent $o_t$ and $p_R$ respectively}\\
2 \ \textbf{if} multiple paths with $k$ nodes\\
3 \ \quad Choose the one with the minimum Euclidean distance\\
4 \ \textbf{end if}\\
5 \ $\mathcal{O}_R = \{o_i | i \mbox{ is the index of a node in } \mathcal{V}_R$\}\\
\codecomment{$\mathcal{O}_R$ is implemented using a queue} \\
6 \ \textbf{return} $\mathcal{O}_R$
}
\end{algorithm}

\begin{algorithm}
\caption{\textsc{BasePlanner}}
\label{alg:base}
{\footnotesize
\vspace{.01in}
\textbf{Input:}
object geometry $\mathcal{O}$, target $o_t$, workspace $\mathcal{W}$, robot kinematics $X$, robot position $p_R$, robot size $r_r$, safety margin $r_s$\\
\textbf{Output:}
Done\\

\vspace{-.1in}
1 \ \ $G(\mathcal{V}, \mathcal{E}) = \mbox{\textsc{GenGraph}}(\mathcal{O}, \mathcal{W}, p_R, r_r, r_s)$\\
2 \ \ $\mathcal{O}_R = $\,\textsc{RelocPath}$(G, o_t, p_R)$\codecomment{compute the initial plan}\\
3 \ \ \textbf{while} $o_t$ is not grasped\\
4 \ \ \quad $o = $\textsc{Dequeue}$(\mathcal{O}_R)$\\
5 \ \ \quad $result =$ \textsc{MotionPlanning}$(o, \mathcal{O}, \mathcal{W}, p_R, X)$\\
6 \ \ \quad \textbf{if} $result$\\
7 \ \ \quad \quad Remove $o$ from $\mathcal{O}$\\
8 \ \ \quad \quad Update $\mathcal{O}$ with sensor inputs\\
9 \ \ \quad \quad $G(\mathcal{V}, \mathcal{E}) = \mbox{\textsc{GenGraph}}(\mathcal{O}, \mathcal{W}, p_R, r_r, r_s)$\\
10 \ \quad \textbf{else}\\
11 \ \quad \quad $G(\mathcal{V}, \mathcal{E}) = G(\mathcal{V}, \mathcal{E} \setminus (v_R, v))$\codecomment{remove the infeasible edge}\\ \codecomment{$(v_R, v)$ representing the path between $p_R$ and $o$}\\
12 \ \quad \textbf{end if}\\
13 \ \quad \textbf{if} a new object(s) is found\codecomment{update the graph and replan}\\
14 \ \quad \quad $G(\mathcal{V}, \mathcal{E}) = \mbox{\textsc{GenGraph}}(\mathcal{O}, \mathcal{W}, p_R, r_r, r_s)$\\
15 \ \quad \textbf{end if}\\
16 \ \quad $\mathcal{O}_R = $\,\textsc{RelocPath}$(G, o_t, p_R)$\\
17 \ \quad \textbf{if} $\mathcal{O}_R = \emptyset$ \textbf{ and } $o_t$ is not grasped\\
18 \ \quad \quad \textbf{return} Fail\codecomment{terminate if no path to $o_t$ is available}\\\codecomment{owing to motion planning failures}\\
19 \ \quad \textbf{end if}\\
20 \ \textbf{end while}\\
21 \ \textbf{return} Done
}\end{algorithm}

Lastly, we propose an online planner that also can deal with Case III together with other cases. In Case III, the robot is tasked with target search first.  In~\cite{dogar2014object}, an object is chosen to be removed such that  the volume revealed after the removal is maximized. In our preliminary study~\cite{nam2019planning_arxiv}, we tested three simple search strategies that are the one similar to \cite{dogar2014object} (which we call \textit{Volume} strategy) and two based on the Euclidean distance between the end-effector and objects. Experiments showed that \textit{Volume} outperforms others so we choose the strategy in this present work. While the target is not detected, Alg.~\ref{alg:reloc} removes object $o$ that reveals the largest volume (lines~1--21). If motion planning for picking and placing $o$ is successful, it is removed and the scene is updated. If no motion is found, the object becomes inaccessible so another object is chosen. If motion planning fails for all objects, Alg.~\ref{alg:reloc} terminates. Once the target is found, the base planner dealing with Cases I and II is executed.\footnote{The pseudocode assumes that the target does not become undetected once it is detected for simplicity. However, it can be modified easily for the case where the target is not recognize even after it is located.}

\begin{algorithm}
\caption{\textsc{RelocPlanner}}
\label{alg:reloc}
{\footnotesize
\vspace{.01in}
\textbf{Input:}
object geometry $\mathcal{O}$, target $o_t$, workspace $\mathcal{W}$, robot kinematics $X$, robot position $p_R$, robot size $r_r$, safety margin $r_s$\\
\textbf{Output:}
Done\\
 
\vspace{-.1in}
1 \ \ \textbf{while} $o_t$ is not detected\\
2 \ \ \quad $G(\mathcal{V}, \mathcal{E}) = \mbox{\textsc{GenGraph}}(\mathcal{O}, \mathcal{W}, p_R, r_r, r_s)$\\
3 \ \ \quad $(\mathcal{O}_A, \mathcal{V}_A)= $\textsc{GetAccessibleObj}$(G)$\\
4 \ \ \quad \textbf{for each} $o_i \in \mathcal{O}_A$\\
5 \ \ \quad \quad Compute the metric $m_i$\codecomment{revealed volume}\\
6 \ \ \quad \textbf{end for}\\
7 \ \ \quad $result = False$\\
8 \ \ \quad \textbf{while} $\neg result$\\
9 \ \ \quad \quad $o = \argmax_{o_i \in \mathcal{O}_A} m_i$\\
10 \ \quad \quad $result =$ \textsc{MotionPlanning}$(o, \mathcal{O}, \mathcal{W}, p_R, X)$\\
11 \ \quad \quad \textbf{if} $result$\\
12 \ \quad \quad \quad Remove $o$ from $\mathcal{O}$\\
13 \ \quad \quad \quad Update $\mathcal{O}$ and $\mathcal{O}_A$ with sensor inputs\\
14 \ \quad \quad \textbf{else}\\
15 \ \quad \quad \quad $\mathcal{O}_A = \mathcal{O}_A \setminus o$  \\
16 \ \quad \quad \quad \textbf{if} $\mathcal{O}_A = \emptyset$\\
17 \ \quad \quad \quad \quad \textbf{return} Fail\codecomment{terminate if motion planning fails for all objects}\\
18 \ \quad \quad \quad \textbf{end if}\\
19 \ \quad \quad \textbf{end if}\\
20  \ \quad \textbf{end while}\\
21 \ \textbf{end while}\\
22 \ \textsc{BasePlanner}$(\mathcal{O}, o_t, \mathcal{W}, X, p_R, r_r, r_s)$\\
23 \ \textbf{return} Done
}\end{algorithm}
\vspace{-2pt}

\begin{figure}[h]
    \vspace{-5pt}
    \centering
    \captionsetup{skip=0pt}
   \begin{subfigure}{0.13\textwidth}
   \centering
   \captionsetup{skip=0pt}
	\includegraphics[width=0.77\textwidth]{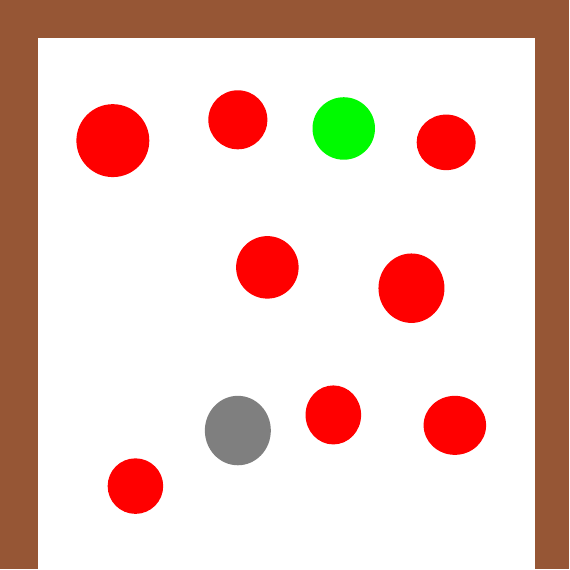}
    \caption{An object configuration}
    \label{fig:test_config}
  \end{subfigure}
  \begin{subfigure}{0.13\textwidth}
  \captionsetup{skip=0pt}
  \centering
	\includegraphics[width=0.87\textwidth]{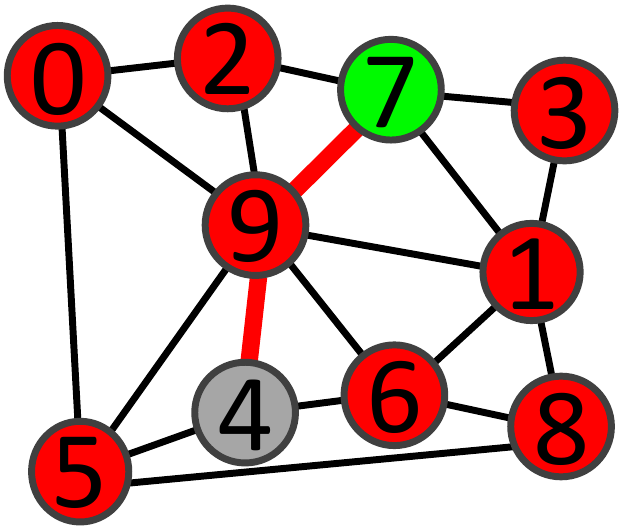}
    \caption{The graph from the configuration}
    \label{fig:test_graph}
  \end{subfigure}
  \begin{subfigure}{0.2\textwidth}
  \captionsetup{skip=0pt}
  \centering
	\includegraphics[width=0.57\textwidth]{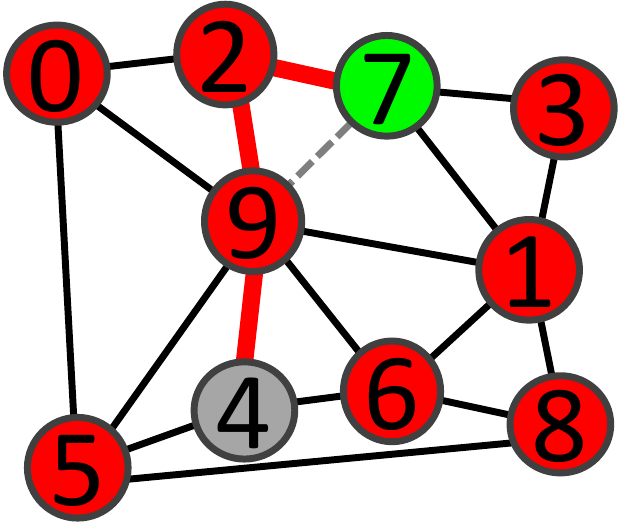}
    \caption{Replanning owing to a motion planning failure}
    \label{fig:test_graph_mp}
  \end{subfigure}
    \caption{An example execution for Case I. (a) The target (green) is surrounded by obstacles (red). The gray circle is the first obstacle to be removed. (b) The red bold edges show the min-hop path. (c) If no motion plan found for grasping the target after removing $o_9$, the edge $(v_9, v_7)$ is removed from $G$ and a new path is found.}
  \label{fig:test}
  \vspace{-15pt}
\end{figure}

\begin{figure}[h]
    \centering
    \captionsetup{skip=0pt}
   \begin{subfigure}{0.23\textwidth}
   \centering
   \captionsetup{skip=0pt}
	\includegraphics[width=0.8\textwidth]{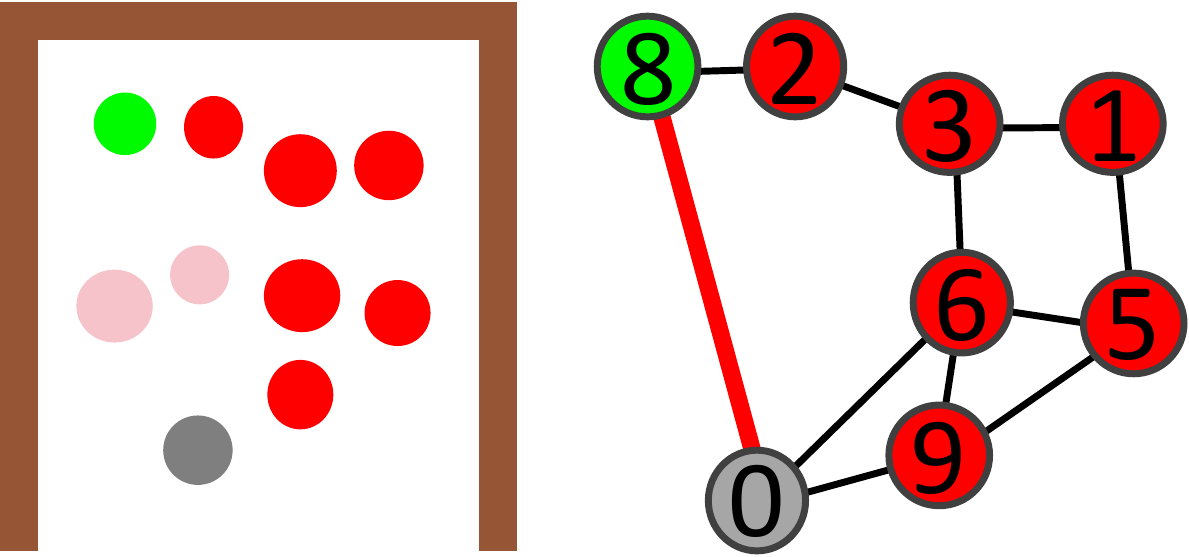}
    \caption{The initial path with two hidden objects (pink)}
    \label{fig:dynamic_ex_a}
  \end{subfigure}\quad
  \begin{subfigure}{0.235\textwidth}
  \centering
  \captionsetup{skip=0pt}
	\includegraphics[width=0.8\textwidth]{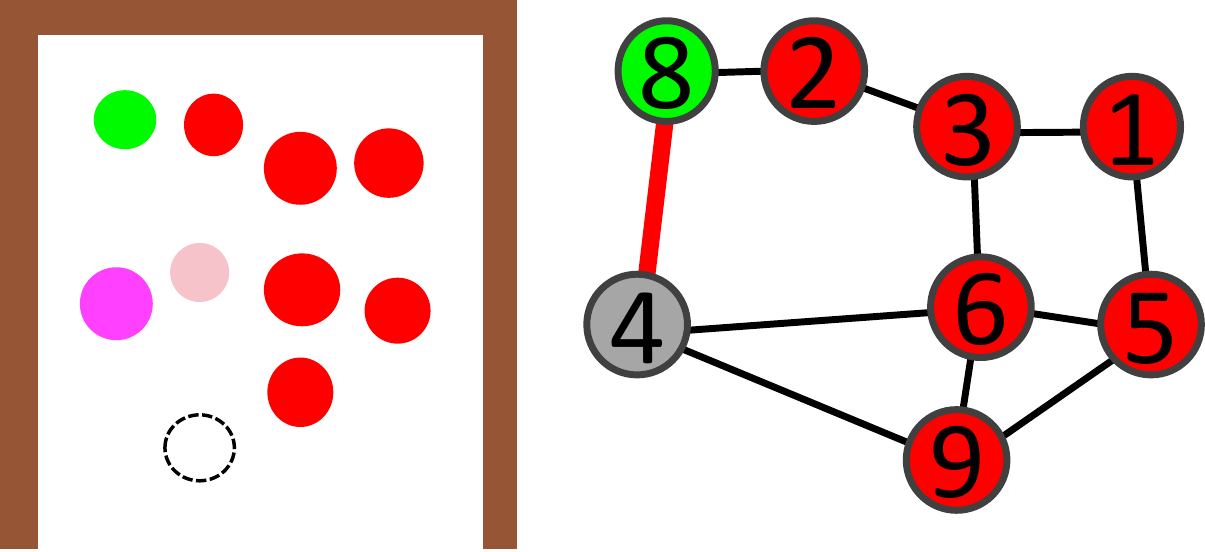}
    \caption{An object revealed (magenta)}
    \label{fig:dynamic_ex_b}
  \end{subfigure}
  \begin{subfigure}{0.235\textwidth}
  \centering
  \captionsetup{skip=0pt}
	\includegraphics[width=0.8\textwidth]{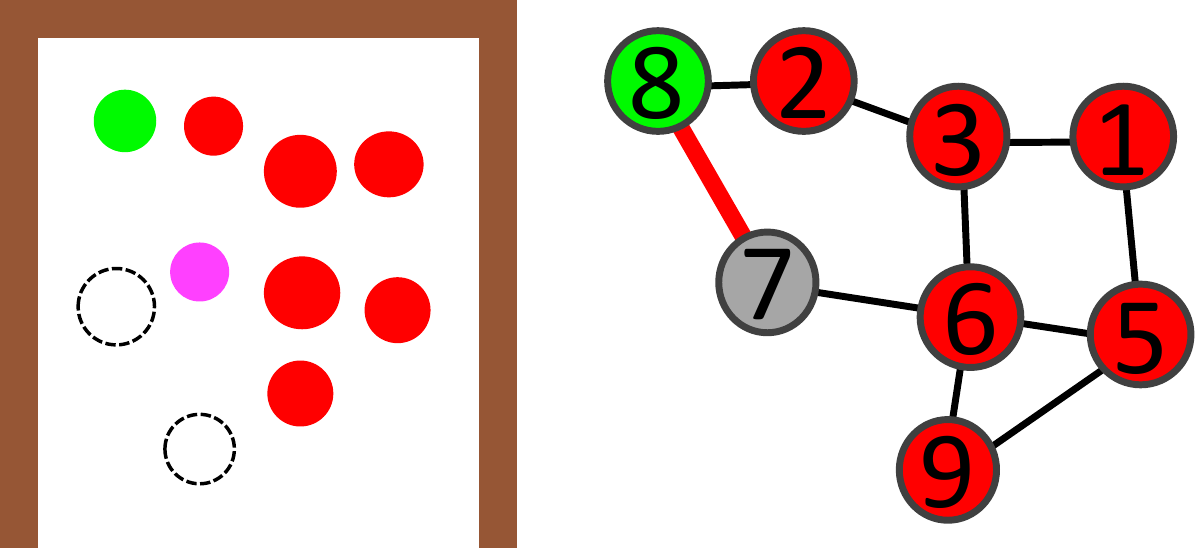}
    \caption{Another object revealed}
    \label{fig:dynamic_ex_c}
  \end{subfigure}\
  \begin{subfigure}{0.235\textwidth}
  \centering
  \captionsetup{skip=0pt}
	\includegraphics[width=0.8\textwidth]{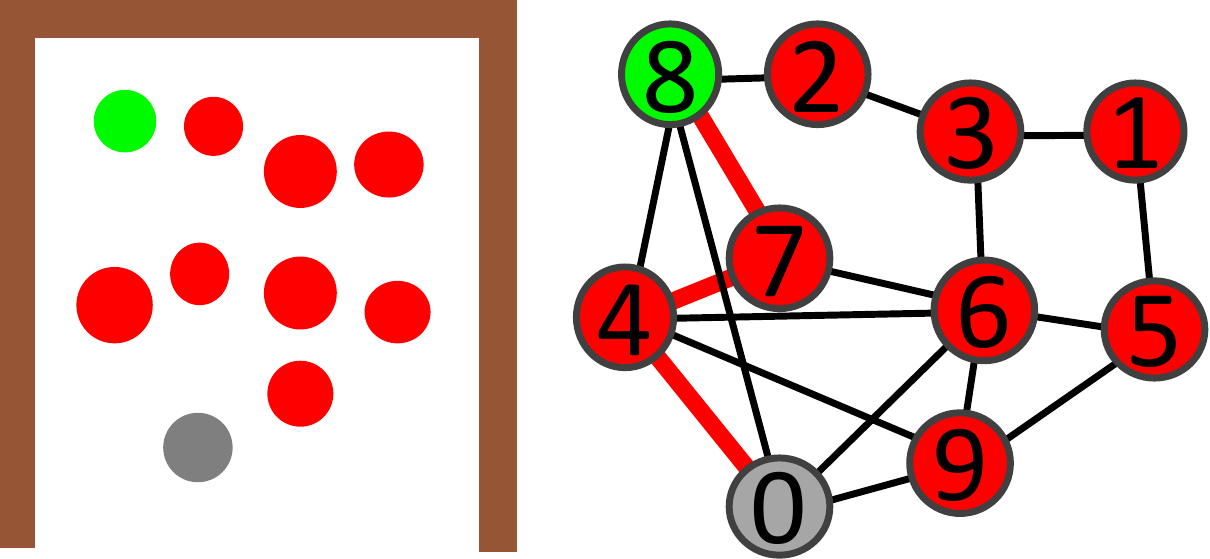}
    \caption{The final path}
    \label{fig:dynamic_ex_final}
  \end{subfigure}
    \caption{An example of Case II. (a) The two pink objects are hidden. (b) After $o_0$ is removed, $o_4$ (in magenta) occurs so a new path is computed. (c) After $o_4$ is removed, $o_7$ is revealed. (d) The final path is shown.}
  \label{fig:dynamic_ex}
  \vspace{-10pt}
\end{figure}

In Fig.~\ref{fig:test}, we show an example result of running Alg.~\ref{alg:reloc} for Case I. Fig.~\ref{fig:dynamic_ex} is an example execution for Case II. Two objects are initially undetected and revealed during execution. In Fig.~\ref{fig:uncertain}, \textit{Volume} strategy removes objects sequentially until the target is detected.

\begin{figure}[h]
    \centering
    \captionsetup{skip=0pt}
   \begin{subfigure}{0.40\textwidth}
   \centering
   \captionsetup{skip=0pt}
	\includegraphics[width=0.8\textwidth]{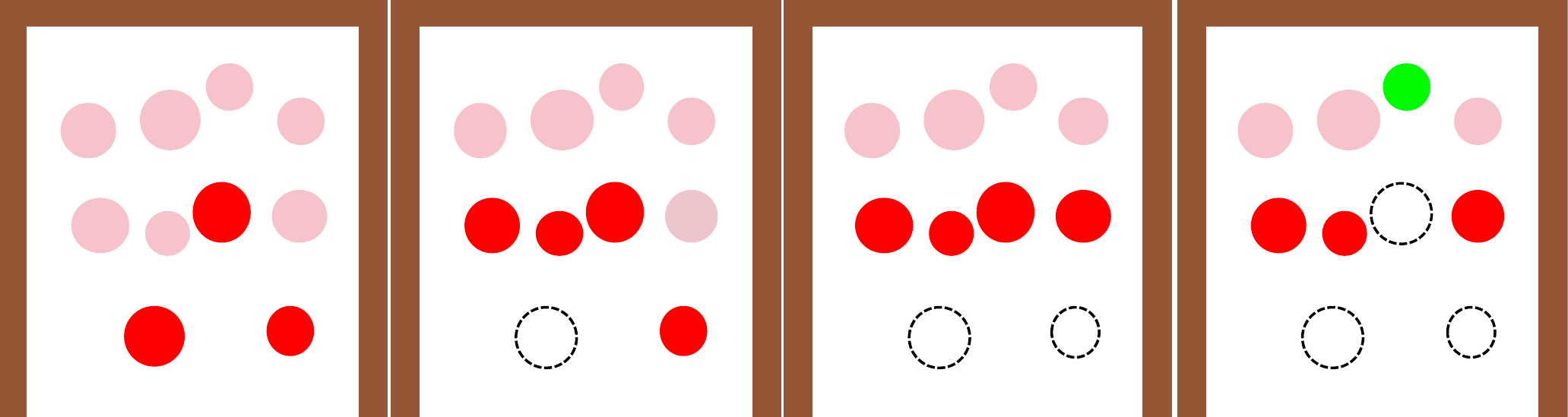}
    \caption{Target search using \textit{Volume} strategy}
    \label{fig:uncertain_ex_progress}
  \end{subfigure}
  \begin{subfigure}{0.49\textwidth}
  \centering
  \captionsetup{skip=0pt}
	\includegraphics[width=0.41\textwidth]{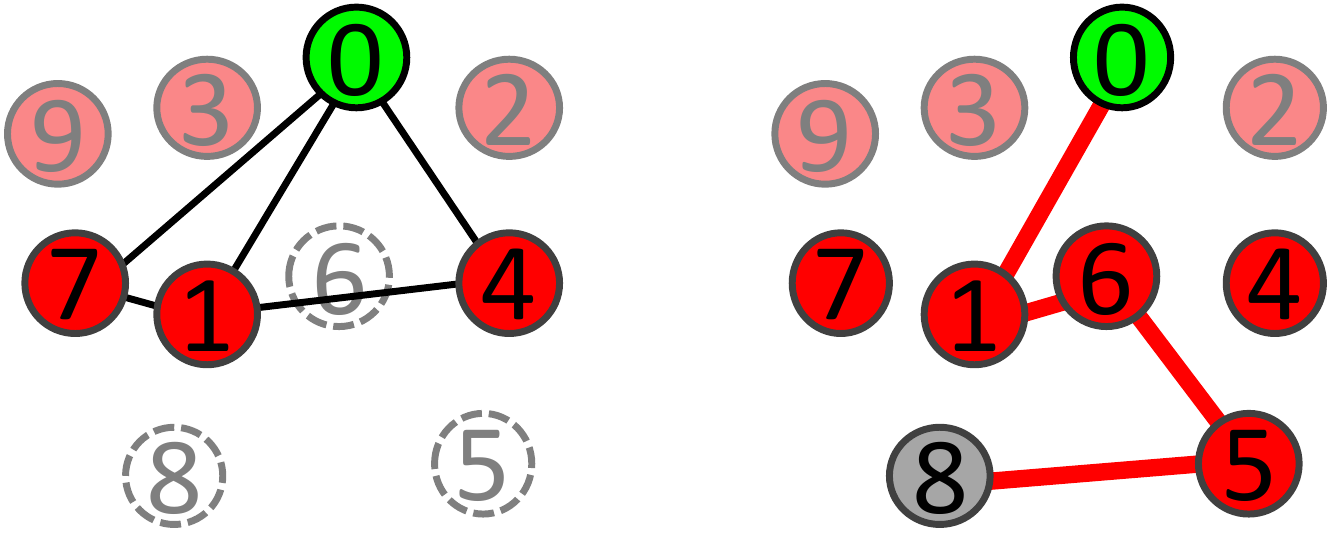}
    \caption{The graph after the target is detected (left) and the final path (right)}
    \label{fig:uncertain_ex_final}
  \end{subfigure}
    \caption{An example of Case III. (a) An object is chosen to be removed in each step such that removing the object maximizes the volume revealed. Until the target is found, objects are relocated and some new objects are discovered. (b) Once the target is found, Alg.~\ref{alg:base} finds a plan.}
  \label{fig:uncertain_ex}
  \vspace{-15pt}
\end{figure}

\subsection{Analysis of algorithms}
\label{sec:analysis}
\vspace{-2pt}
We provide the analysis of some algorithms. 

\lem \textbf{4.1.} Alg.~\ref{alg:graph} has polynomial time complexity.

\pf. Collision checking (line~8) runs for every pair of objects (lines~5--11). There are total $N(N-1)/2$ pairs. The collision checker that we used runs in $O(N(N+N)) = O(N^2)$~\cite{lee2019efficient}. Thus, the time complexity is $O((N(N-1)/2) \times (N(N+N))) = O(N^4)$. \qed
\smallskip

\thm \textbf{4.2.} Alg.~\ref{alg:path} has polynomial time complexity.

\pf. The graph $G(\mathcal{V}, \mathcal{E})$ has $N$ nodes and at most $N(N-1)/2$ edges (fully connected). BFS runs in time $O(|\mathcal{E}| + |\mathcal{V}|) = O(N(N-1)/2 + N+1) = O(N^2)$ to find all min-hop paths between $v_R$ and $v_t$. Choose the one with the shortest Euclidean distance takes $O(N)$. Thus, the time complexity is $O(N^2 + N) = O(N^2)$. \qed
\smallskip

\thm \textbf{4.3.} Alg.~\ref{alg:path} is complete if $G$ is connected\footnote{A graph is connected if there is a path between every pair of nodes. A graph that is not connected has more than one nodes which are completely isolated (so has no edge).}.

\pf. First, we want to show that Alg.~\ref{alg:graph} is complete. Collision checking in line~8 of Alg.~\ref{alg:graph} is complete~\cite{lee2019efficient}, which means that $G$ is constructed after a finite number of iterations.
By definition of connected graphs, a path exists from one of the accessible nodes to the target. BFS is complete~\cite{cormen2009introduction} so finds the shortest path. Thus, Alg.~\ref{alg:path} always returns a path which is the shortest. \qed
\smallskip

Since Algs.~\ref{alg:base} and \ref{alg:reloc} include motion planning, they do not have polynomial time complexities in the number of objects. Their time complexities depend on the motion planner used. Note that sampling-based planners have time complexities that range from $O(n\log n)$ to $O(n^2)$ for both the process and query where $n$ is the number of samples~\cite{karaman2011sampling}.

\section{Experiments}
\label{sec:exp}
\vspace{-2pt}

In this section, we show the experimental results of our method. We measure the task and motion planning time and the success rate of Alg.~\ref{alg:reloc}. We also runs experiments in a simulated environment to measure the total running time including the execution of pick-and-place actions. 
We consider scenarios for all Cases I, II and III. We assume the line of sight of the fixed camera on the robot in Cases II and III. In Case II, 20\% of the total objects are hidden initially and revealed if some front objects are removed. In Case III, only the front objects can be recognized. 

We compare our method with other two fast planning methods. The first one is adopted from~\cite{dogar2012planning} which removes obstacles on the distance-optimal path of the end-effector (\textit{Distance}). The second one is the method presented in~\cite{lee2019efficient} described in Sec.~\ref{sec:related} (\textit{VFH+}). We choose these two methods since ours aims to achieve fast planning for practical uses. While other methods have shown long running time even with small instances as reviewed in Sec.~\ref{sec:related}, the chosen methods often run faster than ours. Since it is not known how they replan if motion planning fails, we implement replanning methods for them for fair comparisons instead of just terminating. We assume that it is beneficial if additional objects are removed when motion planning fails because having a larger configuration space increases the chance to succeed in planning motions. In \textit{Distance}, more objects are added by increasing the width of the distance-optimal path of the end-effector (increasing the width by 2\,cm for every failure from the initial width 6\,cm). In \textit{VFH+}, the angle range for finding obstacles to be removed increases by 10\,deg for every failure from the initial value 90\,deg. 

Motion planning is implemented using a motion planning library~\cite{sucan2012open} in MoveIt motion planning framework~\cite{moveit}. We use RRTConnect~\cite{kuffner2000rrt} which shows the best performance in our pilot studies. The same motion planner is used for all compared methods. We impose a time limit for task and motion planning. A successful planning takes less than one minute. Considering time limits in other work (e.g., 30 mins in~\cite{krontiris2015dealing}), our time limit is significantly short even with the large number of objects (up to 20 objects) and the incorporation of motion planning. The number of objects would not be increased beyond 20 since we use a manipulator with a fixed base so the range the robot can reach is limited. The system is with Intel Core i9 3.6GHz with 32G RAM and Python 2.7.

\subsection{Algorithm tests}
\label{sec:exp_alg}
\vspace{-2pt}
We test the algorithms with random instances to measure the computation time for task and motion planning. We randomly generate 20 instances for each size where $N = 12, 16, 20$. 
We use a model of Kinova JACO1, which is a 6-DOF manipulator in MoveIt. The values used for the robot size $r_r$ and safety margin $r_s$ are 5.0\,cm and 0.5\,cm, respectively. Objects are uniformly distributed at random in a workspace where the dimension is 0.9\,m (length) $\times$ 0.45\,m (width) $\times$ 0.45\,m (height). The diameters and heights of objects are randomly sampled from $\mathcal{U}(5, 6)$ and $\mathcal{U}(6, 7)$ where the unit is centimeter. 

\begin{figure}
\captionsetup{skip=0pt}
    \centering
   \begin{subfigure}{0.235\textwidth}
   \centering
   \captionsetup{skip=0pt}
	\includegraphics[width=0.899\textwidth]{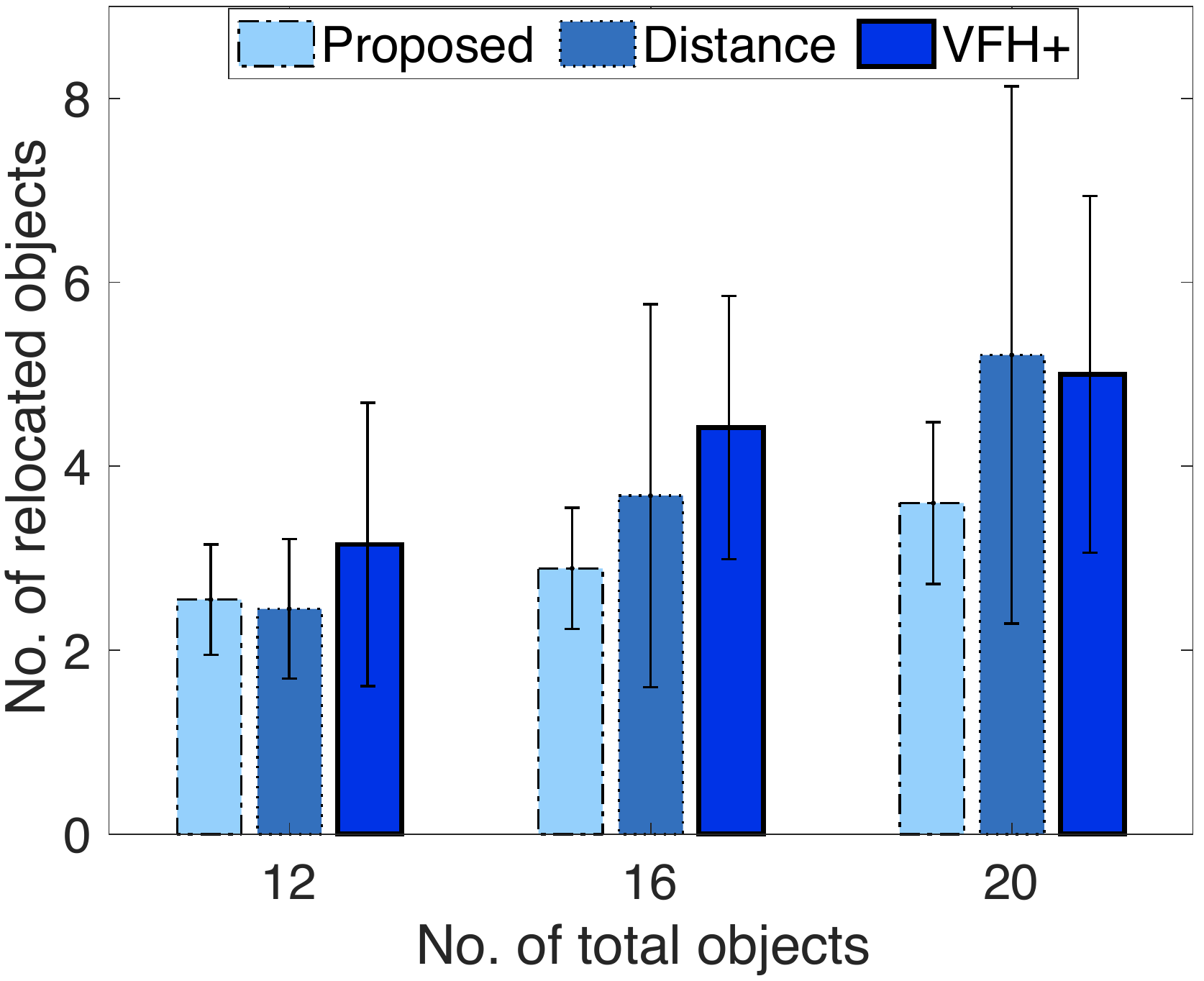}
    \caption{The number of relocated objects}
    \label{fig:sim_num}
  \end{subfigure}%
  \begin{subfigure}{0.22\textwidth}
  \centering
  \captionsetup{skip=0pt}
	\includegraphics[width=\textwidth]{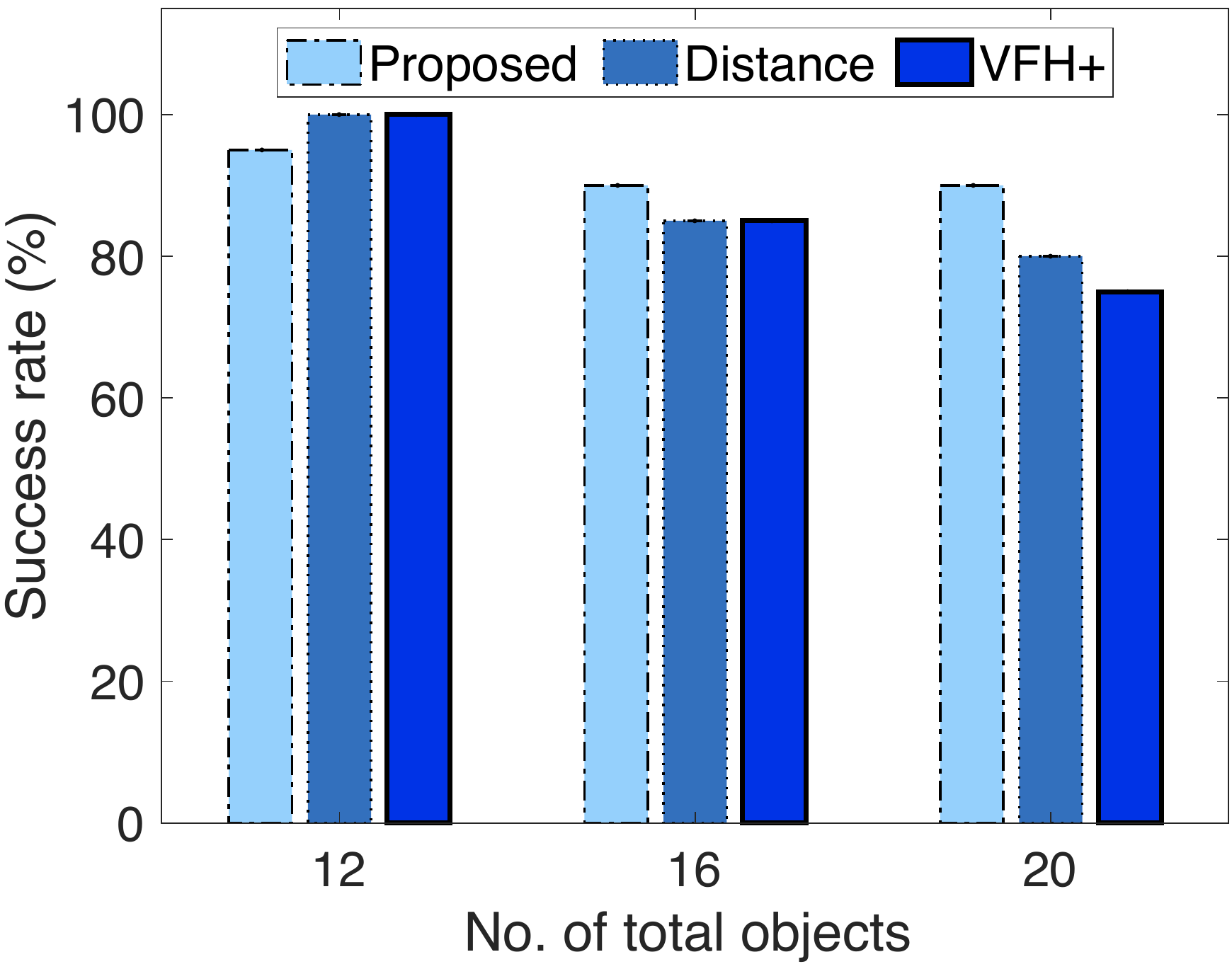}
    \caption{The success rate}
    \label{fig:sim_time}
  \end{subfigure}
    \caption{The comparison between different algorithms for Case I}
  \label{fig:sim}
  \vspace{-12pt}
\end{figure}

\begin{figure}
\vspace{-5pt}
\captionsetup{skip=0pt}
    \centering
   \begin{subfigure}{0.235\textwidth}
   \centering
   \captionsetup{skip=0pt}
	\includegraphics[width=0.899\textwidth]{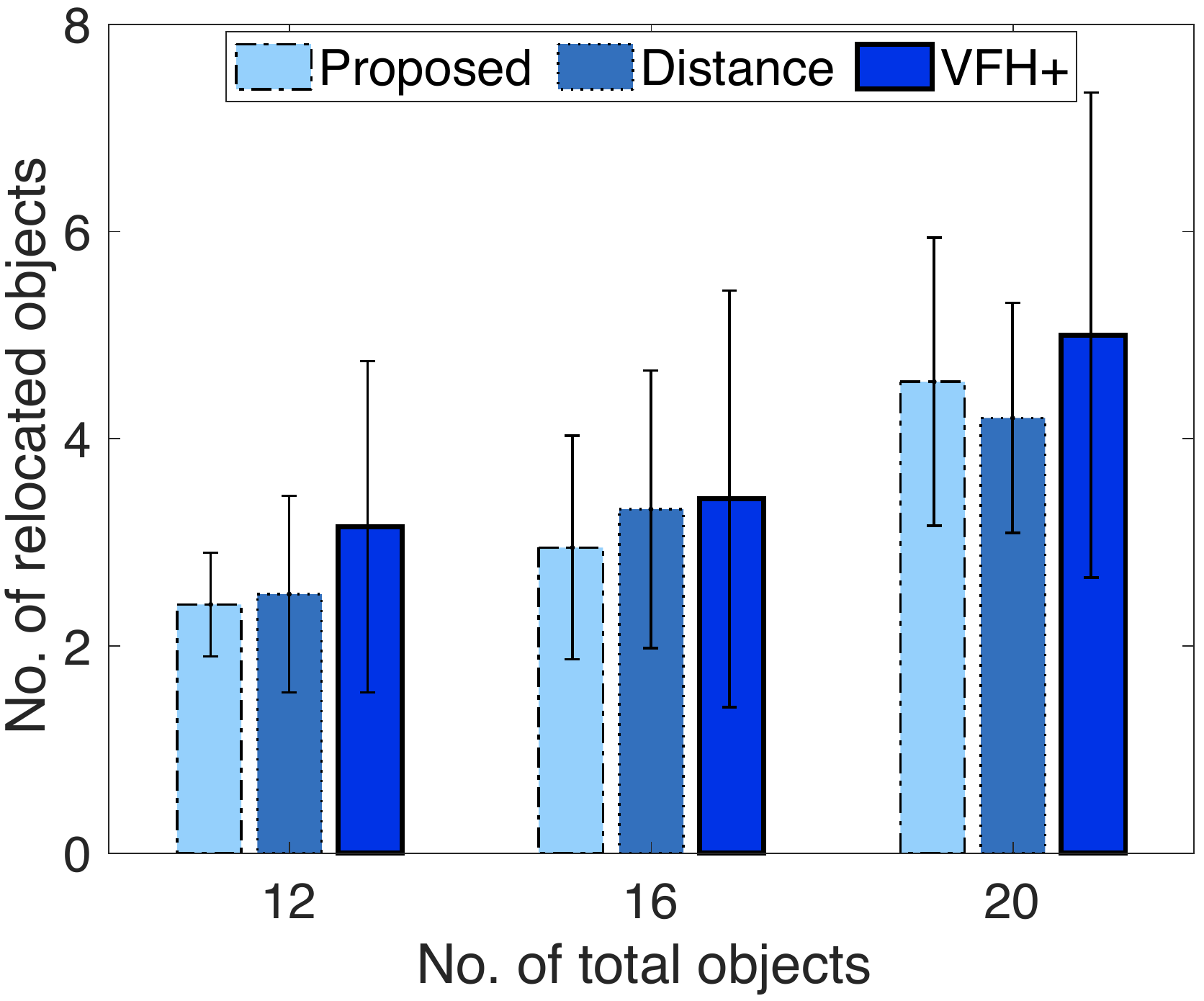}
    \caption{The number of relocated objects}
    \label{fig:sim_num}
  \end{subfigure}%
  \begin{subfigure}{0.22\textwidth}
  \centering
  \captionsetup{skip=0pt}
	\includegraphics[width=\textwidth]{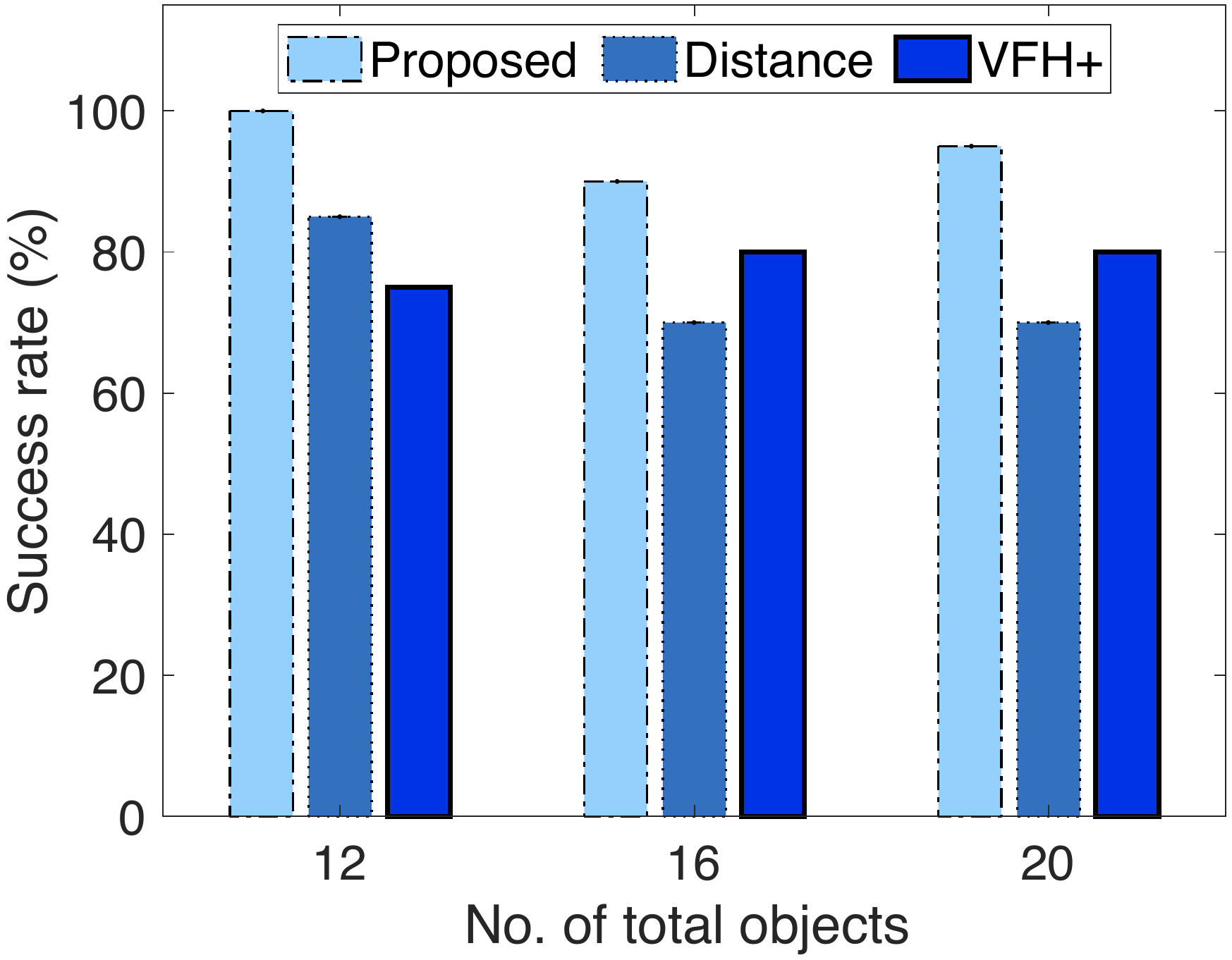}
    \caption{The success rate}
    \label{fig:sim_success}
  \end{subfigure}
    \caption{The comparison between different algorithms for Case II}
  \label{fig:perf}
  \vspace{-5pt}
\end{figure}

We measure the number of relocated objects (i.e., pick-and-place actions) which is the main objective value. We also measure the success rate given the time limit (1 min). We first run 20 instances and compute the success rate. Then we run additional instances to collect 20 data points to have the equal sample size for computing the statistics. Comparisons are done for Cases I and II because the compared methods cannot deal with Case III as they must know the pose of the target. The result is shown in Figs.~\ref{fig:sim}--\ref{fig:perf} and Tables~\ref{tab:perf}--\ref{tab:perf2}.

\begin{table*}[h!]
\captionsetup{skip=0pt}
\caption{Task and motion planning results of the proposed and compared algorithms for Cases I, II, and III (20 repetitions). Except the number of relocated objects and the success rate, all numbers show the mean time (sec) and standard deviation (in parentheses). The number of relocated objects includes the target object.}
\label{tab:perf}
\centering
\scalebox{0.85}{
\begin{tabular}{|c|c||c|c|c||c|c|c||c|c|c|}
\hline
\multirow{2}{*}{Case} & \multirow{2}{*}{Measure} & \multicolumn{3}{c||}{$N = 12$} & \multicolumn{3}{c||}{$N = 16$}& \multicolumn{3}{c|}{$N = 20$}\\
\cline{3-11}
  & & Proposed & Distance & VFH+ & Proposed & Distance & VFH+ &Proposed & Distance & VFH+ \\
\hline
\multirow{4}{*}{I} & \textbf{\#relocated objects} & 2.55 (0.60) & 2.45 (0.76) & 3.15 (1.54) & 2.89 (0.66) & 3.68 (2.08) & 4.42 (1.43) & 3.6 (0.88) & 5.21 (2.92) & 5.00 (1.94) \\
\cline{2-11}
 & \textbf{Success rate} (\%) & 95 & 100 & 100 & 90 & 85 & 85 & 90 & 80 & 75 \\
\cline{2-11}
 & Total time & 13.99 (8.22) & 7.61 (5.06) & 7.62 (3.64) & 23.03 (10.79) & 14.05 (9.68) & 13.21 (8.51) & 27.64 (10.25) & 22.33 (14.23) & 16.85 (8.26) \\
\cline{2-11}
& Time per action & 5.29 (2.06) & 2.97 (1.19) & 2.48 (0.72) & 8.06 (3.31) & 3.94 (2.26) & 2.94 (1.54) & 7.65 (2.22) & 4.14 (1.54) & 3.35 (1.18) \\
\hline
\multirow{4}{*}{II} & \textbf{\#relocated obstacle} & 2.4 (0.50) & 2.5 (0.95) & 3.15 (1.60) & 2.95 (1.08) & 3.32 (1.34) & 3.42 (2.01) & 4.55 (1.39) & 4.2 (1.11) &  5.00 (2.34)\\
\cline{2-11}
 & \textbf{Success rate} (\%) & 100 & 85 & 75 & 90 & 70 & 80 & 95 & 70 & 80 \\
\cline{2-11}
 & Total time &9.45 (3.90)  & 8.03 (7.97) & 10.92 (7.29) & 20.43 (12.26) & 12.29 (7.99) & 11.78 (7.52) & 35.88 (18.17) & 18.76 (14.95) & 21.38 (14.89) \\
\cline{2-11}
& Time per action & 3.86 (1.04) & 2.94 (1.51) & 3.58 (2.06) & 6.72 (2.76) & 3.65 (1.64) & 3.58 (2.06) & 7.67 (2.00) & 4.11 (2.23) & 4.18 (2.05) \\
\hline
\end{tabular}}
\vspace{-8pt}
\end{table*}

\begin{table}
\captionsetup{skip=0pt}
\caption{Task and motion planning results of the proposed method for Case III (20 repetitions). Compared methods cannot solve the case so only our results are shown.}
\label{tab:perf2}
\centering
\scalebox{0.89}{
\begin{tabular}{|c|c|c|c|}
\hline
Measure & $N = 12$ & $N = 16$ & $N = 20$\\
\hline
\textbf{\#relocated obstacle} & 3.25 (1.16) & 4.37 (1.80) & 4.50 (1.47) \\
\hline
\textbf{Success rate} (\%) & 85 & 75  & 70  \\
\hline
Total time & 21.25 (13.97) & 26.09 (15.00) &  32.00 (15.04) \\
\hline
Time per action & 6.39 (3.79) & 5.93 (2.49) & 6.94 (2.23)  \\
\hline
\end{tabular}}
\vspace{-15pt}
\end{table}

\noindent \textit{Discussion: }
Motion planning time has large variances since it depends on the poses of the objects and the configuration space of the robot differing from each random test instance. Motion planning time does not have significant differences across the compared methods. 
The total planning time of ours is longer than others in average. However, ours still takes up to 8\,sec per each action including motion planning in all cases, which is not prohibitively long in such complex environments. The planning time in Case III is much shorter because the method chooses the object to relocate instantly based on the revealed volume but does not plan globally.

The number of relocated objects is reduced 28.0\% compared to VFH+ and 30.9\% compared to Distance in Case I. If only obstacles are counted, the reductions are 35.0\% and 38.2\%, respectively. In Case II, the amount of improvement decreases since hidden objects incur inefficient relocation. The others are less affected by hidden objects as they do not plan globally across the entire workspace. The numbers of relocated objects from the compared methods look better than what they actually are because the instances failed owing to the timeout are not included in computing the mean. Such instances have large numbers of relocated objects while our method rarely has such instances.

Our method has the highest success rate for all instance sizes and the cases. Distance does not incorporate the relationship between objects in task planning thus the order of removing objects is determined without considering what objects should be removed to make next objects reachable. VFH+ shows better success rates than Distance since it considers the relationship. However, it is worse than ours because it is a local planner so may stuck in local minima where no object can be reachable. The success rate in Case III decreases since target search takes long in some instances.

\subsection{Experiments in simulated environments}
\begin{figure}
\vspace{-2pt}
    \centering
   \begin{subfigure}{0.20\textwidth}
    \centering
	\includegraphics[scale=0.12]{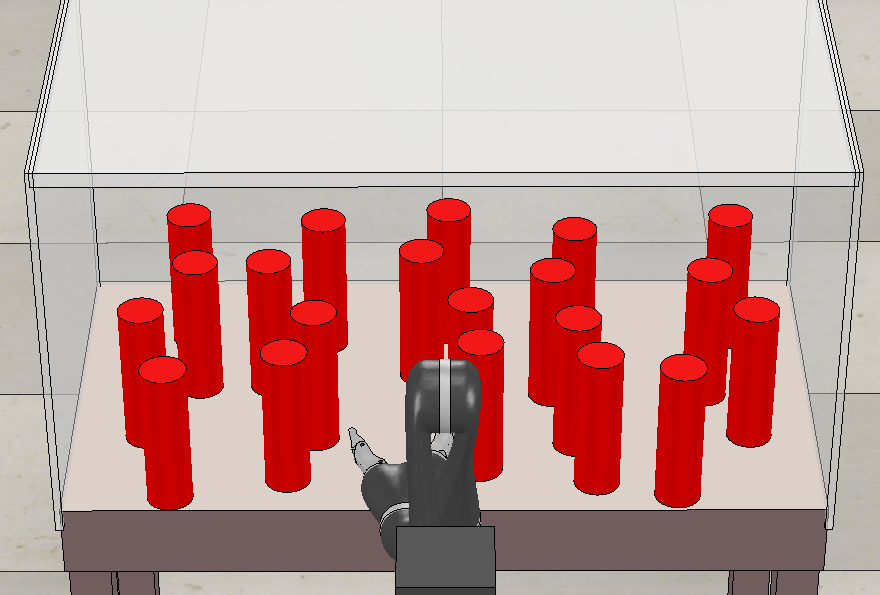}
  \end{subfigure}
  \begin{subfigure}{0.27\textwidth}
    \centering
	\includegraphics[scale=0.137]{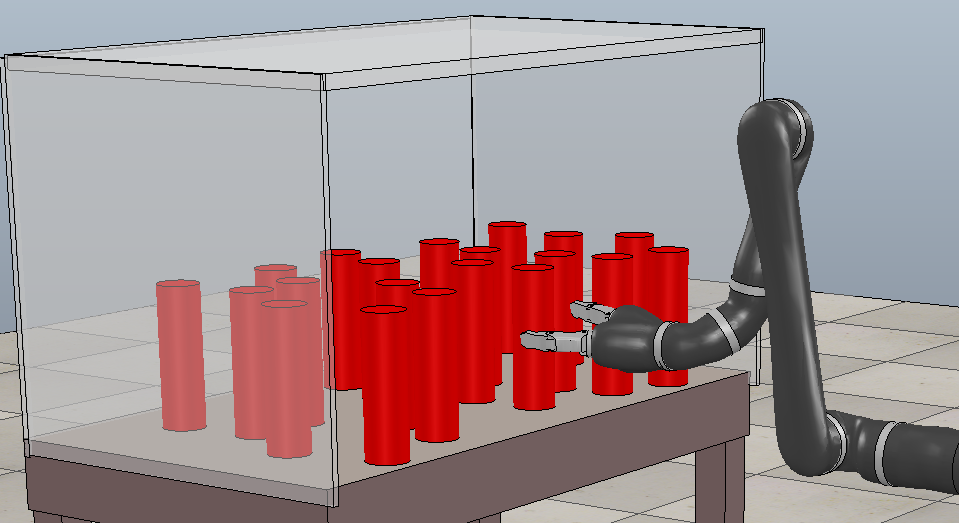}
  \end{subfigure}
    \caption{The simulated environment implemented in V-REP.}
  \label{fig:vrep}
  \vspace{-5pt}
\end{figure}

\vspace{-2pt}
We test the proposed method in a simulated environment using a high fidelity robotic simulator V-REP~\cite{rohmer2013v} with Vortex physics engine. JACO1 is also used in the environment (Fig.~\ref{fig:vrep}). The environment and object specification are the same with the tests in Sec.~\ref{sec:exp_alg}. We compare the three methods in Case I and generate 10 random instances where $N=20$. We omit other experiments owing to the space limit but the results are similar. The robot places removed obstacles in the box below the table. We measure the number of relocated objects, the total running time which includes planning and execution, and the success rate (Table~\ref{tab:sim}).

\noindent \textit{Discussion}: As shown in Sec.~\ref{sec:exp_alg}, the number of relocated objects reduces when our method is used. The reductions of 1.08 and 0.41 in the number of objects reduce 35.0\,sec and 20.4\,sec in total execution time with Distance and VFH+, respectively. The reductions in time are 32.2\% and 21.7\%. The success rate is higher in our method which is similar to the algorithm test in Sec.~\ref{sec:exp_alg}. The result shows that our method can finish the target retrieval task efficiently while maintaining high success rates.

\begin{table}
\captionsetup{skip=0pt}
\caption{The results of simulations in V-REP (10 repetitions) for Case I when $N=20$. The average number of relocated objects and total running time are compared with other methods.}
\label{tab:sim}
\centering
\scalebox{0.89}{
\begin{tabular}{|c|c|c|c|}
\hline
\multirow{2}{*}{Measure} & \multicolumn{3}{c|}{Method} \\
\cline{2-4}
 & Proposed & Distance & VFH+ \\
\hline
Total time (sec) & 73.63 (25.37) & 108.59 (40.49) & 94.04 (29.56) \\
\hline
 \#relocated objects & 2.8 (0.79) & 3.88 (0.99) & 3.21 (0.60) \\
\hline
Success rate (\%) & 100 & 80 & 80 \\
\hline
\end{tabular}}
\vspace{-15pt}
\end{table}

\section{Conclusion}
\vspace{-2pt}
In this work, we study the problem of retrieving objects from clutter. Our objective is to minimize the number of objects to be relocated (pick-and-place actions) to generate a collision-free path for a robotic manipulator so as to reduce the total running time to retrieve the target object. We take the TAMP approach and develop an efficient replanning scheme if motion planning fails. In addition to known environments, we consider partially known environments incurred by occlusions. The results from extensive experiments show that our method reduces the objective value compared to baseline methods. The experiment using a dynamic simulator shows that our approach could works as expected for real problems. In the future, we will consider different shapes of objects so objects may have limited reachable directions. We will also consider non-prehensile actions like pushing and dragging since some objects may need to be moved slightly to avoid collisions.

\bibliographystyle{IEEEtran}
\bibliography{references}

\end{document}